\pdfoutput=1
\documentclass[11pt]{article}

\usepackage{ACL2023}

\usepackage{times}
\usepackage{latexsym}
\usepackage{graphics}
\usepackage{booktabs}
\usepackage{tabularx}

\usepackage{algorithm}
\usepackage{algorithmic}
\usepackage{microtype}      % microtypography

\usepackage{xcolor}         % colors
\usepackage{multirow}
\usepackage{adjustbox}
\usepackage{amsfonts}       % blackboard math symbols
\usepackage{nicefrac} 
\usepackage{amsmath}
\usepackage{newfloat}
\usepackage{listings}
\usepackage{caption} 

\usepackage{todonotes}
\usepackage{float}
\usepackage[T1]{fontenc}
\usepackage[utf8]{inputenc}

\usepackage{microtype}

\usepackage{inconsolata}

\title{QDyLoRA: Quantized Dynamic Low-Rank Adaptation for Efficient Large Language Model Tuning}

\author{
    Hossein Rajabzadeh\textsuperscript{\rm 1}\textsuperscript{\rm 2},
    Mojtaba Valipour \textsuperscript{\rm 1}, 
    Tianshu Zhu \textsuperscript{\rm 2},
    Marzieh Tahaei \textsuperscript{\rm 2},\\
    \textbf{Hyock Ju Kwon} \textsuperscript{\rm 1},  
    \textbf{Ali Ghodsi} \textsuperscript{\rm 1},
    \textbf{Boxing Chen} \textsuperscript{\rm 2},
    \textbf{Mehdi Rezagholizadeh} \textsuperscript{\rm 2} \\
    \textsuperscript{\rm 1} University of Waterloo,
    \textsuperscript{\rm 2} Huawei Noah's Ark Lab\\
    \small{\textmd{\{hossein.rajabzadeh, mojtaba.valipour, hjkwon, ali.ghodsi\}@uwaterloo.ca}} \\
    \small{\textmd{\{mehdi.rezagholizadeh, tianshu.zhu, marzieh.tahaei, boxing.chen\}@huawei.com}} \\
}

\begin{document}
\maketitle
\begin{abstract}
Finetuning large language models requires huge GPU memory, restricting the choice to acquire Larger models. While the quantized version of the Low-Rank Adaptation technique, named QLoRA, significantly alleviates this issue, finding the efficient LoRA rank is still challenging. Moreover, QLoRA is trained on a pre-defined rank and, therefore, cannot be reconfigured for its lower ranks without requiring further fine-tuning steps. This paper proposes QDyLoRA -Quantized Dynamic Low-Rank Adaptation-, as an efficient quantization approach for dynamic low-rank adaptation. Motivated by Dynamic LoRA, QDyLoRA is able to efficiently finetune LLMs on a set of pre-defined LoRA ranks. QDyLoRA enables fine-tuning Falcon-40b for ranks  1 to 64 on a single 32 GB V100-GPU through one round of fine-tuning. Experimental results show that QDyLoRA is competitive to QLoRA and outperforms when employing its optimal rank.
\end{abstract}

\section{Introduction}
%The popularity of adopting Large Language Models (LLMs) in diverse range of down-streaming tasks have been rapidly increasing since last two years. LLMs performs significantly better when being finetuned on down-streaming tasks \cite{chung2022scaling,wei2021finetuned, wang2022super,wang2022self}.   

The popularity of adopting Large Language Models (LLMs) across a diverse range of downstream tasks has rapidly increased over the past two years. Fine-tuning LLMs has become necessary to enhance their performance and introduce desired behaviors while preventing undesired outputs \cite{ding2023parameter}. However, as the size of these models increases, fine-tuning costs become more expensive. This has led to a large body of research that focuses on improving the efficiency of the fine-tuning stage \cite{liu2022few,mao2021unipelt,hu2021lora,edalati2022krona,sung2022lst}. 

Low-rank adapter (LoRA) \cite{hu2021lora} is a well-known, parameter-efficient tuning (PEFT) method that reduces memory requirements during fine-tuning by freezing the base model and updating a small set of trainable parameters in form of low-rank matrix multiplication added to matrices in the base model. However, the memory demand during fine-tuning remains substantial due to the necessity of a backward pass through the frozen base model during stochastic gradient descent.

Recent research has thus focused on further reducing memory usage by designing new parameter-efficient modules that can be tuned without necessitating gradients from the base models \cite{sung2022lst}. Alternatively, researchers have explored combining other efficiency strategies with parameter-efficient tuning methods \cite{kwon2022alphatuning, dettmers2023qlora}.

Among these approaches, QLoRA \cite{dettmers2023qlora} stands out as a recent and highly efficient fine-tuning method that dramatically decreases memory usage. It enables fine-tuning of a 65-billion-parameter model on a single 48GB GPU while maintaining full 16-bit fine-tuning performance. QLoRA achieves this by employing 4-bit NormalFloat (NF4), Double Quantization, and Paged Optimizers as well as LoRA modules.

However, another significant challenge when utilizing LoRA modules is the need to tune their rank as a hyperparameter. Different tasks may require LoRA modules of varying ranks. In fact, it is evident from the experimental results in the LoRA paper that the performance of models varies a lot with different ranks, and there is no clear trend indicating the optimal rank. 
On the other hand, any hyperparameter tuning for finding the optimal rank contradicts the primary objective of efficient tuning and is not feasible for very large models. %Also, the LORA modules when tuned for a fixed rank lack flexibility at inference time. 
Moreover, when deploying a neural network on diverse devices with varying configurations, the use of higher ranks can become problematic for highly sensitive devices due to the increased parameter count. To address this, one typically has to choose between training multiple models tailored to different device configurations or determining the optimal rank for each device and task. However, this process is costly and time-consuming, even when using techniques like LoRA.

\begin{table*}[htb!]
\centering
\caption{A comparison between QLoRA and QDyLoRA on the MMLU benchmark, reporting 5-shot test results for LLMs of varying sizes. QDyLoRA is evaluated on ranks [1,2,4,8,16,32,64] and the best rank is reported in brackets.}
\resizebox{0.8\textwidth}{!}{
\begin{tabular}{ccccccc}
\hline
Dataset & \multicolumn{2}{c}{LLaMA-7b} &  \multicolumn{2}{c}{LLaMA-13b} & \multicolumn{2}{c}{Falcon-40b}  \\[2pt]
\hline
 & QLoRA & QDyLoRA & QLoRA & QDyLoRA & QLoRA & QDyLoRA\\
\hline
Alpaca &38.8 [64]  &39.7 [16] &47.8 [64]  &47.6 [8] &55.2 [64] &57.1 [4]  \\[5pt]

OASST1 &36.6 [64]  &36.8 [16]  &46.4 [64]  &47.2 [8] &56.3 [64] & 56.7 [4] \\[5pt]

Self-Instruct &36.4 [64]  &37.2 [8] &33.3 [64] & 41.6 [4] &51.8 [64] &51.1 [4] \\[5pt]

FLAN-v2 &44.5 [64]  &45.9 [4]   &51.4 [64] &52.1 [8] &58.3 [64]  &60.2 [4] \\[5pt]
\hline
% Add more rows as needed
\end{tabular}}

\label{t1}
\end{table*}

% \todo{is this a valid claim? Maybe it's better to say for multiple tasks finding the optimal rank is hard.}
DyLoRA \cite{valipour2022DyLoRA}, is a recent PEFT method that aims to address theses challenges by employing dynamic Low-Rank Adapter (DyLoRA). Inspired by nested dropout, this method aims to order the representations of the bottleneck at low-rank adapter modules.  Instead of training LoRA blocks with a fixed rank, DyLoRA extends training to encompass a spectrum of ranks in a sorted manner. 
The resulting low-rank PEFT modules not only provide increased flexibility during inference, allowing for the selection of different ranks depending on the context, but also demonstrate superior performance compared to LoRA, all without imposing any additional training time.

%Doing so provides a solution to the problem of dynamically adapting LLM for multiple ranks without exceeding the training budget.Moreover, its adaptive nature during inference allows for flexible deployment according to specific needs.

In this paper, we employ the DyLoRA PEFT method in conjunction with the quantization scheme utilized in the QLoRA work, resulting in  QDyLoRA. QDyLoRA has all the aforementioned benefits of DyLoRA but with significant memory reduction both during training and at inference through 4-bit quantization.
We utilize QDyLoRA for efficient fine-tuning of LLaMA-7b, LLaMA-13b, and Falcon-40b models across ranks ranging from 1 to 64, all on a single 32GB V100 GPU. Once tuned, we determine the optimal rank by inferring the model on the test set. Our results reveal that the optimal rank can be quite low, yet it outperforms QLoRA.

\begin{table*}[htb!]
    \caption{Comparing the performance of QLoRA and QDyLoRA across different evaluation ranks. Both models receives the same training settings. Maximum LoRA rank is set to 64. Falcon-40b is adopted as the base LLM. Exact matching and Bleu-score are used as evaluation measurements for GSM8k and Web-GLM, respectively.}
    \centering
    \resizebox{0.8\textwidth}{!}{
    \begin{tabular}{ccccccccc}
        \hline
        Data &Method & \multicolumn{7}{c}{Rank}\\
        \cline{3-9}
        &&\centering1 & \centering 2 & \centering 4 & \centering 8 & \centering 16 \centering & \centering 32 & 64\\
        \hline
        \centering \multirow{2}{*}{Web-GLM}& QLoRA &19.9 &19.9 &19.9 &33.8 &35.2 &52.7 &54.3\\
        \cline{2-9}
        & QDyLoRA &43.3 &\textbf{56.0} &54.9 &53.3 &53.3 &50.5 &50.2\\
        \hline
        \multirow{2}{*}{GSM8k}&QLoRA &8.9 &8.91 &8.9 &15.1 &20.5 &22.6 &28.1\\
        \cline{2-9}
        &QDyLoRA &21.4 &25.3 &28.2 &\textbf{30.6} &29.8 &28.5 &27.4\\
        \hline        
        \end{tabular}}
    \label{t2}
\end{table*}

\subsection{Related Work}
\paragraph{Low-rank PEFT methods}

These methods aim to fine-tune pre-trained LLMs for specific tasks while minimizing computational and memory resources. Low-rank adaptation techniques were inspired by \cite{aghajanyan2020intrinsic}, demonstrating that pre-trained language models possess a low intrinsic dimension. Since then, several works have explored the incorporation of trainable parameters in the form of low-rank up-projection/down-projection during fine-tuning. In \cite{houlsby2019parameter}, the Adapter module includes a down projection, a non-linear function, an up projection, and a residual connection. These modules are sequentially inserted after the feed-forward network (FFN) or attention blocks.

Additionally, \cite{he2021towards} extends the Adapter concept by introducing trainable modules that run in parallel (PA) with the original pre-trained language-model (PLM) module. As a result of this extension, PA has demonstrated improved performance compared to the original Adapter method. One notable approach among these techniques is LoRA \cite{hu2021lora}, which introduces low-rank up-projection/down-projection into various matrices within a PLM. This method offers efficient inference by seamlessly integrating the adapter module into the original model's weight matrices.

\paragraph{Quantization-aware PEFT methods}
Alpha-Tuning \cite{kwon2022alphatuning}, aims to combine parameter-efficient adaptation and model compression. Alpha-Tuning achieves this by employing post-training quantization, which involves converting the pre-trained language model's full-precision parameters into binary parameters and separate scaling factors. During adaptation, the binary values remain fixed for all tasks, while the scaling factors are fine-tuned for the specific downstream task.

QLoRA \cite{dettmers2023qlora} is a more recent quantization-aware PEFT that combines a low-rank adapter with 4-bit NormalFloat (NF4) quantization and Double Quantization (DQ) of the base model to optimize memory usage. NF4 ensures an optimal distribution of values in quantization bins, simplifying the process when input tensors have a fixed distribution. DQ further reduces memory overhead by quantizing quantization constants. 

To manage memory during gradient checkpointing, QLoRA employs Paged Optimizers, utilizing NVIDIA's unified memory feature for efficient GPU memory management. These techniques collectively enable high-fidelity 4-bit fine-tuning while effectively handling memory constraints.

\paragraph{Dynamic PEFT methods}
DyLoRA paper \cite{valipour2022DyLoRA} introduces a novel approach for training low-rank modules to work effectively across a range of ranks simultaneously, eliminating the need to train separate models for each rank. 

Inspired by the concept of nested dropout, the authors propose a method for organizing the representations within low-rank adapter modules. This approach aims to create dynamic low-rank adapters that can adapt well to various ranks, rather than being fixed to a single rank with a set training budget. This is achieved by dynamically selecting ranks during training, allowing for greater flexibility without the need for extensive rank searching and multiple model training sessions.

\begin{algorithm}[tbh]
 \scriptsize
\caption{QDyLoRA - Training and Inference}
\begin{algorithmic}
\REQUIRE $r\in$ [$r_{min}$,$r_{max}$]; $i$: the number of training iterations; $\alpha$: a scaling factor; $p_B$: probability distribution function for rank selection; $X \in \mathbb{R}^{d \times n}$ : all input features to LoRA; $W_0 \in \mathbb{R}^{m \times d}$ the original frozen pre-trained weight matrix, $W_{dw} \in \mathbb{R}^{r \times d}$; $W_{up} \in \mathbb{R}^{m \times r}$; $Q$: Quantizer; $\mathbb{L}_{\downarrow b}^{DY}$: objective function given truncated weights 

\STATE {\color{blue} Initialization:}
\STATE $W_0^{NF4}=Q(W_0)$ {\color{gray} // Quantize $W_0$ to NF4 \color{black}}

\STATE {\color{blue} Iterations:}
\WHILE {$t < i$}
    \STATE {\color{blue} Forward:}

    \STATE $b\sim p_B(.)$ {\color{gray} // sample a specific rank, during test is given \color{black}} 
    
    \STATE $W_{dw \downarrow b} = W_{dw}$[:$b$,:] {\color{gray} // truncate down-projection matrix \color{black}} 

    \STATE $W_{up \downarrow b} = W_{up}$[:,:$b$]  {\color{gray} // truncate up-projection matrix \color{black}} 

    \STATE $W_0^{DDequant-NF4}=\frac{W_0^{NF4}}{c_2^{FP8}/c_1^{FP32}}$ {\color{gray} // dequantized the chunks of the parameters that are needed  \color{black}}

    \STATE $h = W_0^{DDequant-NF4}X^{BF16}+ \frac{\alpha}{b} W_{up \downarrow b}^{BF16} W_{dw \downarrow b}^{BF16} X^{BF16}$ {\color{gray} // calculate the LoRA output \color{black}} 
    
    \STATE {\color{blue} Backward:}
    \STATE $W_{dw\downarrow b}^{BF16}  \leftarrow W_{dw\downarrow b}^{BF16} - \eta \nabla_{W_{dw\downarrow b}^{BF16}} \mathcal{L}_{\downarrow b}^{\mathcal{DY}}$
    
    \STATE  $W_{up\downarrow b}^{BF16}  \leftarrow W_{up\downarrow b} - \eta \nabla_{W_{up\downarrow b}^{BF16}} \mathcal{L}_{\downarrow b}^{\mathcal{DY}}$
\ENDWHILE
\end{algorithmic}
\label{alg}
\end{algorithm}

\section{Proposed Method: Quantized DyLoRA}

Following DyLoRA notations \cite{valipour2022DyLoRA}, we define a truncated weight $W_{\downarrow b} \in \mathbb{R}^{r \times d}$ as $W[:b,:]$. Assume we have a set of input features $X \in \mathbb{R}^{d\times n}$, a set of pre-trained weights $W_0$, and a given range of desired ranks represented by $r \in$ [$r_{min}$,$r_{max}$] that we want the model to operate with, and a dynamic objective function $L_{\downarrow b}^{DY}$ that 
can evaluate a truncated sub-model. Then we can use the following equation to calculate the forward pass of the model at each iteration.

%$N$ training examples as the pair of $\{x_i\}_{i=1}^N \in X$, and $\{y_i\}_{i=1}^N \in Y$, and we know $Y \subset \mathbb{R}^D$, and $X \subset \mathbb{R}^D$. The goal is to learn a mapping function $f_{\theta}: Y \rightarrow X $ from $Y$ to $X$ with the following objectives in mind:

% \paragraph{Dynamic Learning of Ranks}

% In DyLoRA, we have the following equation to calculate the forward pass:

\begin{multline}
h = W_0^{DDequant-NF4}x^{BF16} \\
+ \frac{\alpha}{b} W_{up \downarrow b}^{BF16} W_{dw \downarrow b}^{BF16} x^{BF16} 
\label{eq: LoRA}
\end{multline}

\noindent where $\alpha$ is the LoRA scalar, and $b$ is the chosen rank by the $p_B(.)$ during training stage.

Following QLoRA \cite{dettmers2023qlora}, we used 4-bit Normal Float (NF4) for storing the double quantized pre-trained weights. As all the computations need to be calculated in BFloat16 precision, DDequant-NF4 will dequantize the stored data. Similar to \cite{dettmers2023qlora}, we have:

\begin{equation}
    W_0^{DDequant-NF4}=\frac{W_0^{NF4}}{c_2^{FP8}/c_1^{FP32}}
\end{equation}

\noindent where $c_1^{FP32}$ and $c_2^{FP8}$ are quantization constants introduced in \cite{dettmers2023qlora}. % defined as $round(\frac{|X^{FP32}|_{max}}{|X^{FP8}|_{max}})$ and $round(\frac{|X^{FP8}|_{max}}{|X^{FP4}|_{max}})$, respectively.  
After this process, we can update the dynamic LoRA parameters using:

\begin{equation}
\begin{split}
    & W_{dw\downarrow b}^{BF16}  \leftarrow W_{dw\downarrow b}^{BF16} - \eta \nabla_{W_{dw\downarrow b}^{BF16}} \mathcal{L}_{\downarrow b}^{\mathcal{DY}} \\
    & W_{up\downarrow b}^{BF16}  \leftarrow W_{up\downarrow b} - \eta \nabla_{W_{up\downarrow b}^{BF16}} \mathcal{L}_{\downarrow b}^{\mathcal{DY}}
\end{split}
\end{equation}

Algorithm \ref{alg} describes the workflow of our proposed QDyLoRA in detail.

% \paragraph{Quantization for Efficiency}

\begin{table*}[htb!]
    \caption{Comparing the performance of DyLoRA, QLoRA and QDyLoRA across different evaluation ranks. all models receives the same training settings. Maximum LoRA rank is set to 64. The results are reported in terms of exact matching.}
    \centering
    \resizebox{0.8\textwidth}{!}{
    \begin{tabular}{ccccccccc}
        \hline
        Data;LLM &Method & \multicolumn{7}{c}{Rank}\\
        \cline{3-9}
        &&\centering1 & \centering 2 & \centering 4 & \centering 8 & \centering 16 \centering & \centering 32 & 64\\
        \hline
        \centering \multirow{3}{*}{GSM8K;LLaMA-7b}& DyLoRA &12.96 &16.91 &17.06 &\textbf{19.94} &18.50 &18.35 &14.94\\
        \cline{2-9}
        & QLoRA &0.0 &0.0 &0.0 &0.0 &0.0 &0.0 &12.66\\
         \cline{2-9}
        & QDyLoRA &12.59 &15.09 &18.50 &16.76 &16.91 &18.65 &14.71\\
        \hline
        \hline
        \multirow{3}{*}{TriviaQA;LLaMA-7b} &DyLoRA &19.27 &23.20 &22.99 &23.32 &23.25 &\textbf{24.12} &22.43\\
        \cline{2-9}
        &QLoRA &0.0 &0.0 &0.0 &0.0 &0.0 &0.0 &15.52\\
        \cline{2-9}
        &QDyLoRA &6.66 &12.49 &17.16 &19.51 &20.09 &21.65 &20.27\\
        \hline
        \hline
        \multirow{3}{*}{GSM8K;LLaMA2-13b} &DyLoRA &OOM &OOM &OOM &OOM &OOM &OOM &OOM\\
        \cline{2-9}
        &QLoRA &0.0 &0.0 &0.0 &0.0 &0.0 &0.0 &21.08\\
        \cline{2-9}
        &QDyLoRA &1.90 &15.01 &22.97 &\textbf{25.55} &24.26 &23.81 &22.08\\
        \hline        
        \end{tabular}}
    \label{t3}
\end{table*}

\section{Experiments and Evaluation}

This section evaluates the efficiency and efficacy of QDyLoRA through several instruct-fine-tuning tasks. The first experiment compares QDyLoRA with QLoRA on Massively Multitask Language Understating (MMLU) benchmark \cite{hendrycks2020measuring}, consisting of more than 50 different tasks, spanning from fundamental mathematics and U.S. history to computer science and law. As shown in Table \ref{t1}\footnote{The same settings as the original QLoRA work are applied here.}, we finetune LLaMA-7b, LLaMA-13b, LLaMA2-13b, and Falcon40b on different datasets, Alpaca \cite{taori2023stanford}, OASST1 \cite{kopf2023openassistant}, Self-Instruct \cite{wang2022self}, and FLAN-v2 \cite{chung2022scaling}, using QLoRA and QDyLoRA techniques. We use the same training budget and maximum LoRA rank \footnote{The maximum LoRA rank is fixed to 64. While QLoRA's rank is always fixed, QDyLoRA can split the training across ranks in range 1 to 64.} for each technique. The results consistently show that QDyLoRA achieves a superior performance by finding the optimal rank. 

The second experiment provides a more in-depth comparison between QLoRA and QDyLoRA. In particular, we fairly finetuned Falcon-40b on WebGLM \cite{liu2023webglm} and GSM8k \cite{cobbe2021training} benchmarks, and compared their test performances across different ranks. As described in Table \ref{t2}, QDyLoRA attains superior performance, notably when employing its optimal ranks (Rank 2 for Web-GLM and Rank 8 for GSM8k). Furthermore, QDyLoRA exhibits consistent superiority over QLoRA, particularly at lower ranks. These findings emphasize the adaptive nature of QDyLoRA in dynamically adjusting its focus during fine-tuning, leading to enhanced efficiency and efficacy compared to its static counterpart, QLoRA. The third experiment compares the performance of DyLoRA, QDyLoRA, and QLoRA on GSM8k and TriviaQA \cite{joshi2017triviaqa} while adopting LLaMA2-13b and LLaMA-7b as LLMs. Table \ref{t3} reports the results. As the table illustrates, for smaller-size models, i.e. LLaMA-7b, DyLoRA and QDyLoRA both perform superior than QLoRA. For larger models, i.e. LLaMA2-13b, DyLoRA fails due to the out-of-memory (OOM) error while QDyLoRA works the best in such situations.

\section{On the semi-sorted behavior of QDyLoRA}
\label{sec:appendixB}
As shown in Table \ref{t2}, QDyLoRA reveals a semi-sorted performance across ranks. We justify this behavior by pointing out the limited finetuning budget. In a limited budget assumption, QDyLoRA updates its lower ranks more frequently than its higher ranks. That is because of the fact that lower ranks are also updated when higher ranks are selected. In other words, lower ranks have more chance to get updated than higher ranks. Hence, lower ranks are more tuned than higher ranks. 

% Moreover, by increasing ranks to higher values, there is a higher chance that the model receives unnecessary finetuning capacity, providing the model with the chance of overfitting and forgetting its pretrained knowledge. 

\section{Conclusion}
QDyLoRA offers an efficient and effective technique for LoRA-based fine-tuning LLMs on downstream tasks. Eliminating the need for fine-tuning multiple models to find the optimal LoRA rank and offering the possibility of fine-tuning larger LLMs are two main advantages of QDyLoRA. The experimental results demonstrated that the optimal rank for QDyLoRA can be surprisingly low, yet it consistently outperforms QLoRA. QDyLoRA provides greater flexibility for deploying LLMs in various contexts and represents a promising step towards making fine-tuning large language models more accessible and efficient.
% \subsection{Style}

\section*{Limitations}
While the 4-bit QDyLoRA exhibits notable performance, it falls short of achieving the performance levels of full precision fine-tuning. One possible solution could be dynamic quantized DyLoRA (DyQDyLoRA), in which the quantization level could also vary during finetuning. In particular, the finetuning strategy can dynamically switch between different quantization levels based on a predefined learning feedback. Additionally, further research is required to investigate the impact of LoRA\textquotesingle s scalar and the range of underlying ranks in QDyLoRA.

% \section*{Acknowledgment}
% We thank Mindspore, which is a new deep learning computing framework, for partial support of this work.

\bibliography{ref}

\begin{thebibliography}{20}
\expandafter\ifx\csname natexlab\endcsname\relax\def\natexlab#1{#1}\fi

\bibitem[{Aghajanyan et~al.(2020)Aghajanyan, Zettlemoyer, and Gupta}]{aghajanyan2020intrinsic}
Armen Aghajanyan, Luke Zettlemoyer, and Sonal Gupta. 2020.
\newblock Intrinsic dimensionality explains the effectiveness of language model fine-tuning.
\newblock \emph{arXiv preprint arXiv:2012.13255}.

\bibitem[{Chung et~al.(2022)Chung, Hou, Longpre, Zoph, Tay, Fedus, Li, Wang, Dehghani, Brahma et~al.}]{chung2022scaling}
Hyung~Won Chung, Le~Hou, Shayne Longpre, Barret Zoph, Yi~Tay, William Fedus, Eric Li, Xuezhi Wang, Mostafa Dehghani, Siddhartha Brahma, et~al. 2022.
\newblock Scaling instruction-finetuned language models.
\newblock \emph{arXiv preprint arXiv:2210.11416}.

\bibitem[{Cobbe et~al.(2021)Cobbe, Kosaraju, Bavarian, Chen, Jun, Kaiser, Plappert, Tworek, Hilton, Nakano et~al.}]{cobbe2021training}
Karl Cobbe, Vineet Kosaraju, Mohammad Bavarian, Mark Chen, Heewoo Jun, Lukasz Kaiser, Matthias Plappert, Jerry Tworek, Jacob Hilton, Reiichiro Nakano, et~al. 2021.
\newblock Training verifiers to solve math word problems.
\newblock \emph{arXiv preprint arXiv:2110.14168}.

\bibitem[{Dettmers et~al.(2023)Dettmers, Pagnoni, Holtzman, and Zettlemoyer}]{dettmers2023qlora}
Tim Dettmers, Artidoro Pagnoni, Ari Holtzman, and Luke Zettlemoyer. 2023.
\newblock Qlora: Efficient finetuning of quantized llms.
\newblock \emph{arXiv preprint arXiv:2305.14314}.

\bibitem[{Ding et~al.(2023)Ding, Qin, Yang, Wei, Yang, Su, Hu, Chen, Chan, Chen et~al.}]{ding2023parameter}
Ning Ding, Yujia Qin, Guang Yang, Fuchao Wei, Zonghan Yang, Yusheng Su, Shengding Hu, Yulin Chen, Chi-Min Chan, Weize Chen, et~al. 2023.
\newblock Parameter-efficient fine-tuning of large-scale pre-trained language models.
\newblock \emph{Nature Machine Intelligence}, 5(3):220--235.

\bibitem[{Edalati et~al.(2022)Edalati, Tahaei, Kobyzev, Nia, Clark, and Rezagholizadeh}]{edalati2022krona}
Ali Edalati, Marzieh Tahaei, Ivan Kobyzev, Vahid~Partovi Nia, James~J Clark, and Mehdi Rezagholizadeh. 2022.
\newblock Krona: Parameter efficient tuning with kronecker adapter.
\newblock \emph{arXiv preprint arXiv:2212.10650}.

\bibitem[{He et~al.(2021)He, Zhou, Ma, Berg-Kirkpatrick, and Neubig}]{he2021towards}
Junxian He, Chunting Zhou, Xuezhe Ma, Taylor Berg-Kirkpatrick, and Graham Neubig. 2021.
\newblock Towards a unified view of parameter-efficient transfer learning.
\newblock \emph{arXiv preprint arXiv:2110.04366}.

\bibitem[{Hendrycks et~al.(2020)Hendrycks, Burns, Basart, Zou, Mazeika, Song, and Steinhardt}]{hendrycks2020measuring}
Dan Hendrycks, Collin Burns, Steven Basart, Andy Zou, Mantas Mazeika, Dawn Song, and Jacob Steinhardt. 2020.
\newblock Measuring massive multitask language understanding.
\newblock \emph{arXiv preprint arXiv:2009.03300}.

\bibitem[{Houlsby et~al.(2019)Houlsby, Giurgiu, Jastrzebski, Morrone, De~Laroussilhe, Gesmundo, Attariyan, and Gelly}]{houlsby2019parameter}
Neil Houlsby, Andrei Giurgiu, Stanislaw Jastrzebski, Bruna Morrone, Quentin De~Laroussilhe, Andrea Gesmundo, Mona Attariyan, and Sylvain Gelly. 2019.
\newblock Parameter-efficient transfer learning for nlp.
\newblock In \emph{International Conference on Machine Learning}, pages 2790--2799. PMLR.

\bibitem[{Hu et~al.(2021)Hu, Shen, Wallis, Allen-Zhu, Li, Wang, Wang, and Chen}]{hu2021lora}
Edward~J Hu, Yelong Shen, Phillip Wallis, Zeyuan Allen-Zhu, Yuanzhi Li, Shean Wang, Lu~Wang, and Weizhu Chen. 2021.
\newblock Lora: Low-rank adaptation of large language models.
\newblock \emph{arXiv preprint arXiv:2106.09685}.

\bibitem[{Joshi et~al.(2017)Joshi, Choi, Weld, and Zettlemoyer}]{joshi2017triviaqa}
Mandar Joshi, Eunsol Choi, Daniel~S Weld, and Luke Zettlemoyer. 2017.
\newblock Triviaqa: A large scale distantly supervised challenge dataset for reading comprehension.
\newblock \emph{arXiv preprint arXiv:1705.03551}.

\bibitem[{K{\"o}pf et~al.(2023)K{\"o}pf, Kilcher, von R{\"u}tte, Anagnostidis, Tam, Stevens, Barhoum, Duc, Stanley, Nagyfi et~al.}]{kopf2023openassistant}
Andreas K{\"o}pf, Yannic Kilcher, Dimitri von R{\"u}tte, Sotiris Anagnostidis, Zhi-Rui Tam, Keith Stevens, Abdullah Barhoum, Nguyen~Minh Duc, Oliver Stanley, Rich{\'a}rd Nagyfi, et~al. 2023.
\newblock Openassistant conversations--democratizing large language model alignment.
\newblock \emph{arXiv preprint arXiv:2304.07327}.

\bibitem[{Kwon et~al.(2022)Kwon, Kim, Bae, Yoo, Kim, Park, Kim, Ha, Sung, and Lee}]{kwon2022alphatuning}
Se~Jung Kwon, Jeonghoon Kim, Jeongin Bae, Kang~Min Yoo, Jin-Hwa Kim, Baeseong Park, Byeongwook Kim, Jung-Woo Ha, Nako Sung, and Dongsoo Lee. 2022.
\newblock Alphatuning: Quantization-aware parameter-efficient adaptation of large-scale pre-trained language models.
\newblock \emph{arXiv preprint arXiv:2210.03858}.

\bibitem[{Liu et~al.(2022)Liu, Tam, Muqeeth, Mohta, Huang, Bansal, and Raffel}]{liu2022few}
Haokun Liu, Derek Tam, Mohammed Muqeeth, Jay Mohta, Tenghao Huang, Mohit Bansal, and Colin~A Raffel. 2022.
\newblock Few-shot parameter-efficient fine-tuning is better and cheaper than in-context learning.
\newblock \emph{Advances in Neural Information Processing Systems}, 35:1950--1965.

\bibitem[{Liu et~al.(2023)Liu, Lai, Yu, Xu, Zeng, Du, Zhang, Dong, and Tang}]{liu2023webglm}
Xiao Liu, Hanyu Lai, Hao Yu, Yifan Xu, Aohan Zeng, Zhengxiao Du, Peng Zhang, Yuxiao Dong, and Jie Tang. 2023.
\newblock Webglm: Towards an efficient web-enhanced question answering system with human preferences.
\newblock \emph{arXiv preprint arXiv:2306.07906}.

\bibitem[{Mao et~al.(2021)Mao, Mathias, Hou, Almahairi, Ma, Han, Yih, and Khabsa}]{mao2021unipelt}
Yuning Mao, Lambert Mathias, Rui Hou, Amjad Almahairi, Hao Ma, Jiawei Han, Wen-tau Yih, and Madian Khabsa. 2021.
\newblock Unipelt: A unified framework for parameter-efficient language model tuning.
\newblock \emph{arXiv preprint arXiv:2110.07577}.

\bibitem[{Sung et~al.(2022)Sung, Cho, and Bansal}]{sung2022lst}
Yi-Lin Sung, Jaemin Cho, and Mohit Bansal. 2022.
\newblock Lst: Ladder side-tuning for parameter and memory efficient transfer learning.
\newblock \emph{Advances in Neural Information Processing Systems}, 35:12991--13005.

\bibitem[{Taori et~al.(2023)Taori, Gulrajani, Zhang, Dubois, Li, Guestrin, Liang, and Hashimoto}]{taori2023stanford}
Rohan Taori, Ishaan Gulrajani, Tianyi Zhang, Yann Dubois, Xuechen Li, Carlos Guestrin, Percy Liang, and Tatsunori~B Hashimoto. 2023.
\newblock Stanford alpaca: An instruction-following llama model.

\bibitem[{Valipour et~al.(2022)Valipour, Rezagholizadeh, Kobyzev, and Ghodsi}]{valipour2022DyLoRA}
Mojtaba Valipour, Mehdi Rezagholizadeh, Ivan Kobyzev, and Ali Ghodsi. 2022.
\newblock Dylora: Parameter efficient tuning of pre-trained models using dynamic search-free low-rank adaptation.
\newblock \emph{arXiv preprint arXiv:2210.07558}.

\bibitem[{Wang et~al.(2022)Wang, Kordi, Mishra, Liu, Smith, Khashabi, and Hajishirzi}]{wang2022self}
Yizhong Wang, Yeganeh Kordi, Swaroop Mishra, Alisa Liu, Noah~A Smith, Daniel Khashabi, and Hannaneh Hajishirzi. 2022.
\newblock Self-instruct: Aligning language model with self generated instructions.
\newblock \emph{arXiv preprint arXiv:2212.10560}.

\end{thebibliography}
\bibliographystyle{acl_natbib}
% \bibliographystyle{aaai24}

% \clearpage
% \newpage

%%%%%%%%%%%%%%%%%%%%%%%%%%%%%%%%%%%%%%%%%%%%%%%%%%%%%%%%%%%%
\appendix
\section{Supplementary Material}

\subsection{Hyperparameters}
\label{ap:hyperparameters}
Table \ref{tab:hyperparamters-table} provides an overview of the hyperparameters and experimental configurations employed in this study, which are crucial configurations that determine various aspects of the training process and model behavior in this study. Common key parameters across the experiments include the choice of optimizer, Adam-Beta2 value, maximum gradient norm, and warmup ratio, which collectively influence how the model adjusts its weights during training. LoRA-specific parameters such as LoRA dropout probability, maximum LoRA rank, and alpha value control the behavior of LoRA layers. Additionally, double quantization and quantization type impact the precision of numerical representations within the model, which are considered the same as baselines. Learning rate scheduling and weight decay contribute to the optimization process, helping to prevent overfitting and stabilize training. Random seeds ensure reproducibility, while the specified GPU determines the hardware used for training. Each model configuration, whether for the Web-GLM, GSM8k, or the specific experiment outlined in Table 1, features parameters tailored to the characteristics of the dataset and the computational resources available. These hyperparameters collectively shape the training process, ultimately influencing the performance and effectiveness of the models in the study.

\begin{table}[hbt!]
\centering
\resizebox{\columnwidth}{!}{  
\begin{tabular}{l|cc}
\hline
\textbf{Model} & \textbf{Parameter} & \textbf{Value} \\
\hline
\multirow{17}{*}{Common settings} & &  \\
& Optimizer & paged-adamw-32bit \\
& Adam-Beta2 & 0.999 \\
& Max-Grad-Norm & 0.3 \\
& Warmup Ratio & 0.03 \\
& LoRA-Dropout & 0.05 \\
& LoRA-r & 64 \\
& LoRA-Alpha & 16 \\
& LoRA-Module & all \\
& Double-Quant & True \\
& Quant-Type & nf4 \\
& LR Scheduler & Constant \\
& Learning Rate (LR) & 1e-5 \\
& Weight Decay & 0.0 \\
& Seeds & [42]\\
& GPU & Tesla V100-PCIE-32GB \\
\hline
\multirow{6}{*}{Web-GLM} & &  \\
& Max-Steps & 2700 \\
& Gradient-Accumulation-Step & 2 \\
& Source-Max-Len & 2048\\
& Target-Max-Len & 2048 \\
& Batch Size & 16 \\
\hline
\multirow{6}{*}{GSM8k} & &  \\
&  Max-Steps & 300 \\
& Gradient-Accumulation-Step & 2 \\
& Source-Max-Len & 2048\\
& Target-Max-Len & 2048 \\
& Batch Size & 16 \\
\hline
\multirow{6}{*}{Table \ref{t1}} & &  \\
&  Max-Steps & 5000 \\
& Gradient-Accumulation-Step & 16 \\
& Source-Max-Len & 16\\
& Target-Max-Len & 512 \\
& Batch Size & 32 \\
\hline
\end{tabular}}
\caption{
The list of hyperparameters employed across various experiments in our study. The common settings are applied to each experiments. If a parameter is not explicitly mentioned, it means that we use default values from HuggingFace Transformers v'4.28.0.dev0' (source: https://huggingface.co/docs/transformers).
}
\label{tab:hyperparamters-table}
\end{table}

\subsection{Generated Text Quality}

To describe the quality of texts generated by QDyLoRA, a sample query taken from GSM8k is fed to Falcon-40b trained by QDyLoRA. Below, we report the generated answers for different LoRA ranks.

\textbf{Query}: Janet's ducks lay 16 eggs per day. She eats three for breakfast every morning and bakes muffins for her friends every day with four. She sells the remainder at the farmers' market daily for \$2 per fresh duck egg. How much in dollars does she make every day at the farmers' market?

\textbf{Rank=64}: Janet has 16 - 3 - 4 = <<16-3-4=7>>7 eggs left. She sells 7 - 16 = <<7-16=9>>9 eggs at the farmers' market.She makes 9 * $2 = $<<9*2=18>>18 at the farmers' market every day.

\textbf{Rank=32}: She eats 3 eggs for breakfast and bakes 4 muffins per day, so she uses 3 + 4 = <<3+4=7>>7 eggs per day. She has 16 - 7 = <<16-7=9>>9 eggs left over each day. She sells these eggs at the farmers' market for \$2 per egg, so she makes 2 * 9 = <<2*9=18>>18 per day at the market.

\textbf{Rank=28}: She eats 3 eggs for breakfast and bakes 4 muffins per day, so she uses 3+4 = <<3+4=7>>7 eggs per day. She sells 16-7 = <<16-7=9>>9 eggs per day for \$2 per egg. She makes 9*2 = <<9*2=18>>18 per day at the market.

\textbf{Rank=16}: Janet's ducks lay 16 eggs per day and she eats 3 for breakfast every morning so she has 16-3=13 eggs left. Janet bakes 4 muffins every day and she uses 4 eggs so she has 13-4=9 eggs left. Janet sells the remaining eggs at the farmers' market for \$2 each so she makes 9*2=<<9*2=18>>18 dollars a day. Janet makes 18 dollars a day at the farmers' market.

\end{document}